\documentclass[9pt,conference]{IEEEtran}
\IEEEoverridecommandlockouts

\usepackage{cite}
\usepackage{amsmath,amssymb,amsfonts}
\usepackage{algorithmic}
\usepackage{graphicx}
\usepackage{textcomp}
\usepackage{xcolor}

\usepackage{amsthm}
\usepackage{booktabs}
\usepackage{enumitem}
\usepackage{color}
\usepackage{makecell}
\usepackage{utfsym}
\usepackage{multirow}
\usepackage{colortbl}
\usepackage{url}

\def\BibTeX{{\rm B\kern-.05em{\sc i\kern-.025em b}\kern-.08em
    T\kern-.1667em\lower.7ex\hbox{E}\kern-.125emX}}
\begin{document}

\title{
Content and Salient Semantics Collaboration \\ for Cloth-Changing Person Re-Identification
\thanks{This work was supported in part by the National Natural Science Foundation of China under Grant 62176061 and Grant 62406252, in part by Shanghai Technology Development and Entrepreneurship Platform for Neuromorphic and AI SoC, and in part by the Shanghai Research and Innovation Functional Program under Grant 17DZ2260900.}
}

\DeclareRobustCommand*{\IEEEauthorrefmark}[1]{
    \raisebox{0pt}[0pt][0pt]{\textsuperscript{\footnotesize\ensuremath{#1}}}}
    
\author{
\IEEEauthorblockN{
Qizao Wang\IEEEauthorrefmark{1}, Xuelin Qian\IEEEauthorrefmark{2}$^{,\dagger}$, Bin Li\IEEEauthorrefmark{1}, Lifeng Chen\IEEEauthorrefmark{1}, Yanwei Fu\IEEEauthorrefmark{3}, Xiangyang Xue\IEEEauthorrefmark{1}$^{,\dagger}$\thanks{$\dagger$ Co-corresponding authors.}
}
\IEEEauthorblockA{
\IEEEauthorrefmark{1}\textit{School of Computer Science, Fudan University, Shanghai, China} \\
\IEEEauthorrefmark{2}\textit{School of Automation, Northwestern Polytechnical University, Xi'an, China} \\
\IEEEauthorrefmark{3}\textit{School of Data Science, Fudan University, Shanghai, China} \\
qzwang22@m.fudan.edu.cn, xlqian@nwpu.edu.cn, \{libin, chenlf, yanweifu, xyxue\}@fudan.edu.cn
}
}

\maketitle

\begin{abstract}
Cloth-changing person re-identification aims at recognizing the same person with clothing changes across non-overlapping cameras. Advanced methods either resort to identity-related auxiliary modalities (\textit{e.g.}, sketches, silhouettes, and keypoints) or clothing labels to mitigate the impact of clothes. However, relying on unpractical and inflexible auxiliary modalities or annotations limits their real-world applicability. In this paper, we promote cloth-changing person re-identification by leveraging abundant semantics present within pedestrian images, without the need for any auxiliaries. Specifically, we first propose a unified Semantics Mining and Refinement (SMR) module to extract robust identity-related content and salient semantics, mitigating interference from clothing appearances effectively. We further propose the Content and Salient Semantics Collaboration (CSSC) framework to collaborate and leverage various semantics, facilitating cross-parallel semantic interaction and refinement. Our proposed method achieves state-of-the-art performance on three cloth-changing benchmarks, demonstrating its superiority over advanced competitors. The code is available at \url{https://github.com/QizaoWang/CSSC-CCReID}.
\end{abstract}


\section{Introduction}
\label{sec:intro}
Person Re-IDentification (Re-ID) aims to recognize individuals across different cameras and times. With the growing demand for surveillance applications and the resurgence of deep learning, significant efforts have been devoted to the advancement of person Re-ID~\cite{qian2017multi,hou2019interaction,sun2018beyond,wang2023rethinking,meng2024unleashing,wang2024distribution}.
Existing person Re-ID models are principally studied in the short-term scenario~\cite{wang2024large}, where the clothing of the same person remains consistent. Consequently, the learned features heavily rely on clothing appearances, rendering the models ineffective when individuals change their clothes or wear similar clothes as others. As a result, there is a rising interest in addressing the cloth-changing challenge in long-term real-world scenarios~\cite{qian2020long,yang2019person,wang2022co,wang2024image}.

\begin{figure}[t]
\centering
  \includegraphics[width=0.8\linewidth]{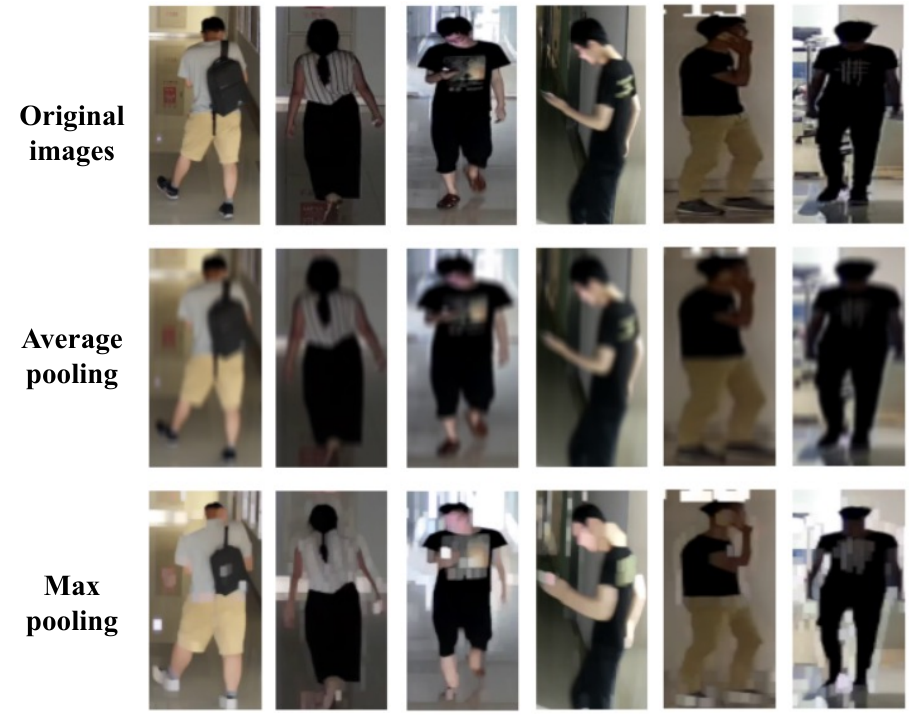}
  \caption{\textbf{Comparison of applying different poolings on pedestrian images.} Average pooling can better preserve image content while smoothing details, while max pooling is better at capturing salient information in the content. Incorporating both of them is expected to learn abundant semantics, and thus improve the discriminative ability of Re-ID models.
  }
  \vspace{-0.1in}
  \label{fig:intro}
\end{figure}

\begin{figure*}[t!]
\centering
  \includegraphics[width=0.75\linewidth]{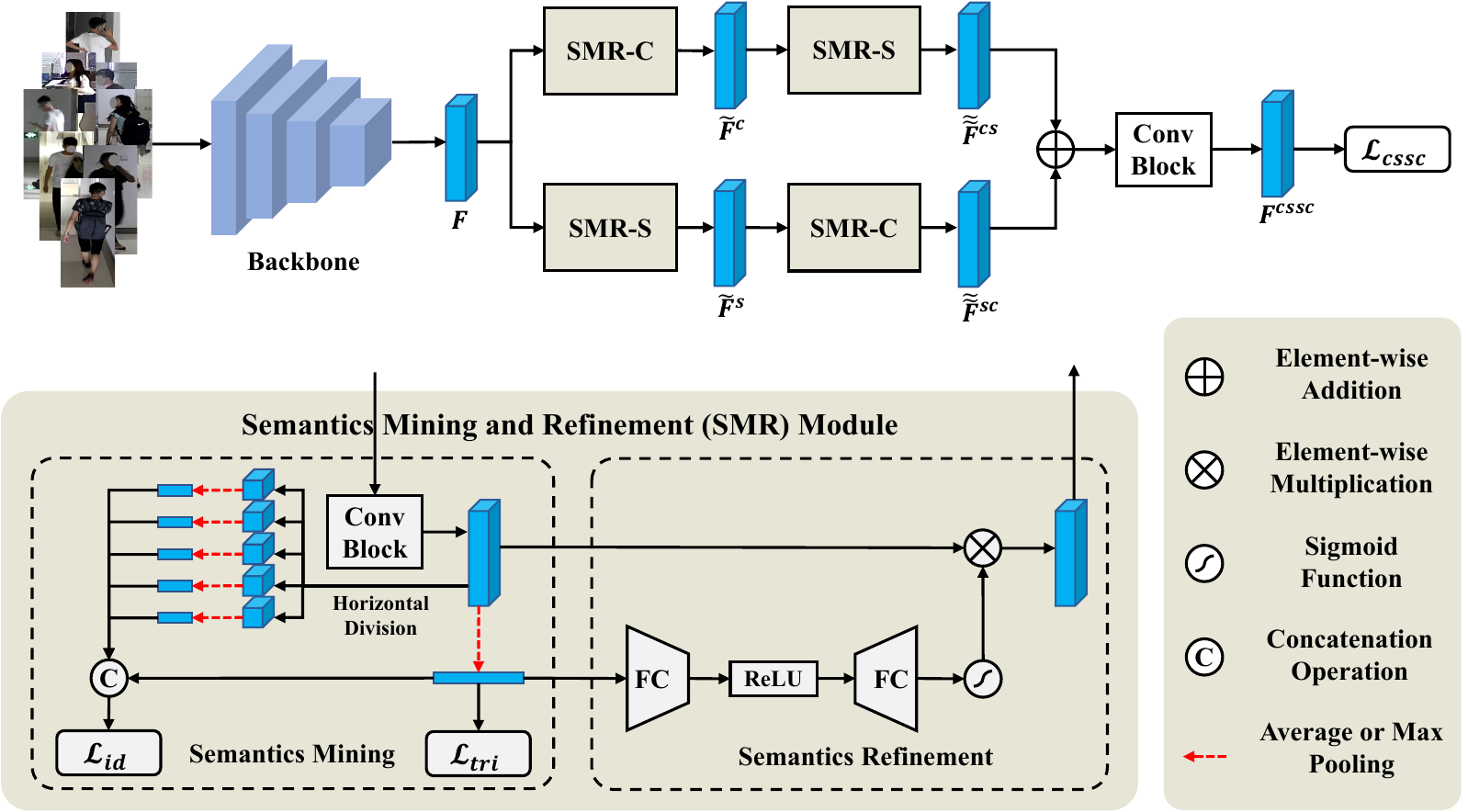}
  \caption{\textbf{Framework of our method.} The Semantics Mining and Refinement (SMR) module learns identity-related semantics without relying on any auxiliaries. SMR modules equipped with average pooling (SMR-C) and max pooling (SMR-S) learn content and salient semantics effectively, respectively. Taking advantage of SMR, our framework interacts and refines both semantics sequentially and parallelly to promote cloth-changing person Re-ID.
  }
  \vspace{-0.1in}
  \label{fig:framework}
\end{figure*}

In this paper, we address the cloth-changing challenge without relying on any auxiliary module~\cite{qian2020long,yang2019person,wang2022co,hong2021fine,jin2022cloth} or extra annotation~\cite{gu2022clothes,yang2023good,han2023clothing}.
Interestingly, we observe that two commonly used pooling strategies, namely average and max poolings, exhibit distinct characteristics in information integration. As depicted in Fig.~\ref{fig:intro}, average pooling effectively preserves image content information while smoothing out some details, but max pooling excels at capturing salient information within the content. 
Motivated by these findings, we expect to leverage them to enhance the model's ability in content understanding and salient information capture, which are both crucial for effective cloth-changing person Re-ID. 
On the one hand, the person Re-ID model is required to comprehend the content of pedestrian images and perceive clothing information to extract identity characteristics adaptively to tackle cloth-changing cases. On the other hand, salient identity cues play a vital role in mitigating confusion from varying clothes and stabilizing learning features robust to clothing variations. 
However, as shown in Fig.~\ref{fig:intro}, employing pooling in the pixel space of original images is susceptible to interference from some identity-irrelevant information, such as background and illumination, which are unrelated to pedestrian identity. 

To this end, we exploit both pooling strategies to facilitate feature learning in the high-level semantic space with the guidance of identity. 
Specifically, we propose the Semantics Mining and Refinement (SMR) module. It guides the model to perceive and extract identity-related semantics, including pedestrian body shape, faces, \textit{etc}. Then, the learned semantic features facilitate model learning adaptively.
To avoid interference from identity-irrelevant information and improve the robustness of the Re-ID model in the cloth-changing scenario, we propose to mine local semantics and guide the mining process with carefully designed identity supervision. 
By integrating the SMR module into the backbone seamlessly, we present the Content and Salient Semantics Collaboration (CSSC) framework capable of sequential and parallel semantic interaction and refinement. Both content and salient semantics play pivotal roles in recognizing and distinguishing pedestrians with interference from varying clothing appearances.

\noindent \textbf{Contributions.} We summarize key contributions as follows.

(1) We propose the novel Content and Salient Semantics Collaboration (CSSC) framework incorporating effective cross-parallel semantic interaction. It effectively addresses the cloth-changing challenge in person Re-ID without relying on unpractical and inflexible auxiliary modalities or extra annotations.

(2) We propose the Semantics Mining and Refinement (SMR) module, which applies the distinct pooling characteristics to learn and utilize robust identity-related content and salient semantics in a unified manner.

(3) Extensive experiments on three cloth-changing person Re-ID benchmarks demonstrate the effectiveness and superiority of our proposed method.

\section{Methodology}
\label{sec:method}

\subsection{Problem Formulation}
Given a training dataset $\mathcal{D} = \{ x_{i}, y_{i}\}_{i=1}^{N}$ containing totally $N$ images and $N^{p}$ identities, where $x_{i}$ and $y_{i}$ represent the $i$-th images and its corresponding identity label, the Re-ID model $\mathcal{G}$ aims to map the person image $x_{i}$ into a discriminative feature representation $F_{i} = \mathcal{G}\left(x_{i}\right)$. 
Subsequently, a pooling layer is then applied to $F_{i} \in \mathbb{R}^{h \times w \times d}$, where $h$ and $w$ are the height and width of the feature representation, yielding the feature vector $f_{i} \in \mathbb{R}^{d}$. 
The model $\mathcal{G}$ is required to tolerate clothing changes and extract the robust feature $f_{i}$, so that the similarity between features of pedestrians with the same identity is larger than that of pedestrians with different identities, regardless of whether they change clothes or not. 
In the subsequent discussion, unless otherwise specified, we omit the subscript $i$ of $F_{i}$ and $f_{i}$ to simplify notation.

\subsection{Semantics Mining and Refinement Module}
To enhance the discriminative ability of the model in the cloth-changing scenario, we leverage both identity-related content and salient semantics with our designed Semantics Mining and Refinement (SMR) module. In the following, we first illustrate the learning of identity-related content semantics which help comprehend pedestrian images and extract identity characteristics adaptively to tackle cloth-changing cases.

\noindent \textbf{Global content semantics mining.} 
With $F$ extracted by the backbone, one convolutional block is appended, resulting in $F^{c} \in \mathbb{R}^{h \times w \times d}$.
Then, we obtain the average pooled feature,
\begin{equation}
f^{c} = {\rm AvgPool}(F^{c}) \in \mathbb{R}^{d},
\label{eq:avg_pool}
\end{equation}
where ${\rm AvgPool}$ denotes the global average pooling operation, which effectively integrates content semantics in feature maps. 

\noindent \textbf{Local content semantics mining.} 
In addition to mining global content semantics, the semantics of local regions are also crucial. Therefore, we horizontally divide $F^{c}$ into different parts and use one convolutional layer for dimensionality reduction, resulting in several local content semantic features ${\{ f^{{l}^{c}_{p}} \}}^{P}_{p=1}$.
However, in the cloth-changing scenario, the reliability of local regions is questionable. Due to clothing changes and clothing similarity between pedestrians, if carried out as previous works~\cite{sun2018beyond,wang2018learning}, local semantics can mislead model learning.
To leverage local content semantics effectively, we concatenate the global and local content semantic features. Formally,
\begin{equation}
\hat{f}^{c} = [f^{c}; f^{{l}^{c}_{1}}; f^{{l}^{c}_{2}}; \cdots; f^{{l}^{c}_{P}}],
\label{eq:cat}
\end{equation}
where $[\cdot; \cdot]$ denotes concatenation in the feature dimension. 
To ensure the mined content semantics are identity-related and discriminative, widely-used Re-ID losses are used as follows:
\begin{equation}
\mathcal{L}_{smr\_c}(F^{c}) = \mathcal{L}_{id}(\hat{f}^{c}) + \mathcal{L}_{tri}(f^{c}),
\label{eq:smr_loss}
\end{equation}
where $\mathcal{L}_{id}$ and $\mathcal{L}_{tri}$ denote widely used identity classification loss~\cite{luo2019bag} and triplet loss~\cite{hermans2017defense}, respectively. Note that $\mathcal{L}_{id}$ includes an identity classifier implemented as one fully connected layer. Additonally, the identity label of $\hat{f}^{c}$ (\textit{i.e.}, $y$) is required in $\mathcal{L}_{id}$ and $\mathcal{L}_{tri}$ for identity supervision, and we omit it for notation simplicity.

\noindent \textbf{Content semantics refinement.} 
To improve the discriminative ability of the model, the mined identity-related global content semantics are also used to refine $F^{c}$ in an adaptive manner. $F^{c}$ is recalibrated along the channel dimension using the learned semantics in $f^{c}$. Formally,
\begin{equation}
\widetilde{F}^{c} = \sigma(W_{2} \phi(W_{1}f^{c})) \otimes F^{c},
\label{eq:se}
\end{equation}
where $\phi$ denotes the \textit{ReLU} activation function, $\sigma$ denotes the \textit{Sigmoid} function, and $\otimes$ denotes element-wise multiplication. 
$W_1 \in \mathbb{R}^{\frac{d}{r} \times d}$ and $W_2 \in \mathbb{R}^{d \times \frac{d}{r}}$ are the weights of two Fully-Connected (FC) layers, where the first one compresses the channel dimension to achieve information bottleneck. $r$ is set to 16 following \cite{hu2018squeeze}.
The final \textit{Sigmoid} function acts as a self-attention mechanism across channels, generating scores to refine the content semantics of pedestrian features adaptively. The $\sim$ notation represents our proposed semantics mining and refinement operation applied to $F$.

\noindent \textbf{Salient semantics mining and refinement.}
Salient identity cues are helpful to mitigate confusion from varying clothes and stabilize learning features robust to clothing variation. 
Following a similar design for learning identity-related content semantics, identity-related salient semantics can be mined with our SMR module by simply replacing average pooling with max pooling. We denote the extracted salient semantic feature as $F^{s}$, and its max pooled feature $f^{s}$ and locally augmented feature $\hat{f}^{s}$ are supervised by the loss $\mathcal{L}_{smr\_s}(F^{s})$ (similar to Eq.~\ref{eq:smr_loss}) for identity-related salient semantics mining. Finally, refined salient semantic feature $\widetilde{F}^{s}$ can be obtained. 

\subsection{Semantics Collaboration Framework}
The SMR modules with average pooling and max pooling, denoted as SMR-C and SMR-S, are used for content and salient semantics mining and refinement, respectively.
As illustrated in Fig.~\ref{fig:framework}, SMR-C and SMR-S are alternatively employed in different orders within two branches.
On the one hand, the refined content semantics $\widetilde{F}^{c}$ facilitate mining salient semantics and result in $\widetilde{\widetilde{F}}\vphantom{F}^{cs}$. The losses involved in the first branch are formulated as follows:
\begin{equation}
\mathcal{L}_{branch1} = \mathcal{L}_{smr\_c}(F^{c}) + \mathcal{L}_{smr\_s}(\widetilde{F}^{cs}).
\label{eq:branch1_loss}
\end{equation}

On the other hand, the refined salient semantics $\widetilde{F}^{s}$ aid in mining content semantics and result in $\widetilde{\widetilde{F}}\vphantom{F}^{sc}$. The losses involved in the second branch are formulated as follows:
\begin{equation}
\mathcal{L}_{branch2} = \mathcal{L}_{smr\_s}(F^{s}) + \mathcal{L}_{smr\_c}(\widetilde{F}^{sc}).
\label{eq:branch2_loss}
\end{equation}

The cross-parallel interaction framework takes full advantage of mined content and salient semantics, promoting the robustness of Re-ID models in the cloth-changing scenario.

\noindent \textbf{Comprehensive semantics learning.} 
The two features from the two branches specialized in content and salient semantics are fused to a comprehensive semantics feature as follows:
\begin{equation}
F^{cssc} = {\rm Conv}(\widetilde{\widetilde{F}}\vphantom{F}^{cs} \oplus \widetilde{\widetilde{F}}\vphantom{F}^{sc}),
\label{eq:fuse}
\end{equation}
where $\oplus$ denotes element-wise addition, and ${\rm Conv}$ denotes a convolutional block. We take the max-pooled $F^{cssc}$ as the final Re-ID identity representation, which is supervised by commonly used Re-ID losses. Formally,
\begin{equation}
f^{cssc} = {\rm MaxPool}(F^{cssc}) \in \mathbb{R}^{d},
\end{equation}
\begin{equation}
\mathcal{L}_{cssc} = \mathcal{L}_{id}(f^{cssc}) + \mathcal{L}_{tri}(f^{cssc}),
\label{eq:fuse_loss}
\end{equation}
where ${\rm MaxPool}$ denotes the global max pooling operation.

\subsection{Training and Inference}
Both content and salient semantics are mined and refined with the help of our proposed SMR module, and effectively collaborate to promote person Re-ID in our proposed CSSC framework. The overall loss is computed as follows:
\begin{equation}
\mathcal{L} = \mathcal{L}_{branch1} + \mathcal{L}_{branch2} + \mathcal{L}_{cssc}.
\label{eq:overall_loss}
\end{equation}

In the cloth-changing scenario, the appearance of pedestrians can vary significantly due to different clothing. 
Introducing $\mathcal{L}_{tri}$ prematurely when the Re-ID model is not robust to clothing variations can lead to the overfitting problem and contribute to suboptimal performance. Therefore, we empirically find it better to introduce $\mathcal{L}_{tri}$ for model optimization after the first learning rate decay. During inference, the comprehensive semantics feature vector $f^{cssc}$ is used to compute the cosine distance between person images for retrieval.

\section{Experiments}
\label{sec:experiment}
\subsection{Experimental Settings}
\noindent \textbf{Datasets.}
To demonstrate the effectiveness of our method, we evaluate it on three widely-used cloth-changing person Re-ID datasets, \textit{i.e.}, PRCC~\cite{yang2019person}, LTCC~\cite{qian2020long}, and Celeb-reID~\cite{huang2019beyond}. 

\noindent \textbf{Implementation details.}
Following previous works, we adopt ResNet-50~\cite{resnet} pre-trained on ImageNet~\cite{deng2009imagenet} and apply our SMR modules with $P=8$ in the \textit{conv5} layer by reusing the three ready-made convolutional blocks. 
Following~\cite{qian2020long,gu2022clothes,cui2023dcr,yan2022weakening,yang2023good}, the input images are resized to $384 \times 192$. The batch size is set to 32. Random horizontal flipping, padding, random cropping, and random erasing~\cite{zhong2020random} are used for data augmentation. Adam optimizer~\cite{kingma2014adam} with weight decay of $5 \times 10^{-4}$ is adopted for 120 epochs. The learning rate linearly increases from $3 \times 10^{-5}$ to $3 \times 10^{-4}$ in the first 10 epochs and decreases by a factor of 10 at the 30th and 60th epochs. All experiments are conducted on one NVIDIA GeForce GTX 1080 Ti with 11GB of memory.

\noindent \textbf{Evaluation metrics.} 
For evaluation, we adopt standard metrics as in most person Re-ID literature, namely Cumulative Matching Characteristic (CMC) curves and mean Average Precision (mAP). For LTCC and PRCC, we evaluate our method under both the standard setting and the cloth-changing setting following~\cite{wang2024exploring}.

\begin{table}[tp]
\centering
\caption{\scriptsize \textbf{Comparison of our method with state-of-the-art methods on PRCC and LTCC.} 
Methods in the gray region use extra ground-truth clothing labels for training. ``sketch'', ``sil.'', and ``pose'' represent contour sketches, silhouettes, and human poses, respectively. Methods marked with ``$\ast$'' involve multiple training stages for extra auxiliary networks. ``Standard'' and ``Cloth-Changing'' mean the standard and cloth-changing settings, respectively. The best results are shown in bold. \label{tab:prcc_ltcc}}
\resizebox{1\linewidth}{!}{
\begin{tabular}{llcccccccc}
\toprule
  \multirow{3}{*}{\textbf{Methods}} &
  \multirow{3}{*}{\textbf{Modality}} &
  \multicolumn{4}{c}{\textbf{PRCC}} &
  \multicolumn{4}{c}{\textbf{LTCC}} \\ \cmidrule(r){3-6} \cmidrule(r){7-10}
  & & \multicolumn{2}{c}{Cloth-Changing} &
  \multicolumn{2}{c}{Standard} &
  \multicolumn{2}{c}{Cloth-Changing} &
  \multicolumn{2}{c}{Standard} \\ \cmidrule(r){3-4} \cmidrule(r){5-6} \cmidrule(r){7-8} \cmidrule(r){9-10}
  & & Rank-1 & {mAP} & Rank-1 & mAP  & Rank-1 & {mAP} & Rank-1 & mAP  \\ \midrule
  
\rowcolor{gray!10} UCAD~\cite{yan2022weakening} & RGB+sil. & 45.3 & {-} & 96.5 & - & 32.5 & {15.1} & 74.4 & 34.8 \\
\rowcolor{gray!10} CAL~\cite{gu2022clothes} & RGB & 55.2 & {55.8} & 100 & 99.8 & 40.1 & {18.0} & 74.2 & 40.8 \\ 
\rowcolor{gray!10} DCR-ReID~\cite{cui2023dcr} & RGB+sil.+sketch & 57.2 & 57.4 & 100 & 99.7 & 41.1 & 20.4 & 76.1 & 42.3 \\
\rowcolor{gray!10} AIM~\cite{yang2023good} & RGB & 57.9 & {58.3} & 100 & 99.9 & 40.6 & {19.1} &  76.3 & 41.1 \\
\rowcolor{gray!10} CCFA$^{\ast}$~\cite{han2023clothing} & RGB & 61.2 & {58.4} & 99.6 & 98.7 & 45.3 & {22.1} & 75.8 & 42.5 \\ \midrule

HA-CNN~\cite{li2018harmonious} & RGB & 21.8 & {-} & 82.5 & - & 21.6 & {9.3} & 60.2 & 26.7 \\
PCB~\cite{sun2018beyond} & RGB & 41.8 & {38.7} & 99.8   & 97.0 & 23.5 & {10.0} & 65.1 & 30.6 \\
IANet~\cite{hou2019interaction} & RGB & 46.3 & {45.9} & 99.4 & 98.3 & 25.0 & {12.6} & 63.7 & 31.0 \\
TransReID~\cite{he2021transreid} & RGB & 46.6  & {44.8} & \bf 100 & {99.0} & 34.4 & {17.1}  & 70.4 & 37.0  \\
RCSANet$^{\ast}$~\cite{huang2021clothing} & RGB & 50.2 & {48.6} & \bf 100 & 97.2 & - & {-} & - & -    \\
ACID~\cite{yang2023win} & RGB & 55.4 & \bf 66.1 & 99.1 & 99.0 & 29.1 & 14.5 & 65.1 & 30.6 \\
FSAM~\cite{hong2021fine} & RGB+pose+sil. & 54.5 & {-} & 98.8 & - & 38.5 & {16.2} & 73.2 & 35.4 \\
GI-ReID$^{\ast}$~\cite{jin2022cloth} & RGB+sil. & 33.3 & {-} & 80.0  & -  & 23.7 & {10.4} & 63.2   & 29.4 \\
CAMC~\cite{wang2022co} & RGB+pose & - & {-} & - & {-} & 36.0 & {15.4} & 73.2 & 35.3 \\
\midrule

CSSC (Ours) & RGB & \textbf{65.5} & {63.0} & \textbf{100} & \textbf{99.1} & \textbf{43.6} & {\textbf{18.6}} &  \textbf{78.1} &  \textbf{40.2} \\ 
\bottomrule
\end{tabular}}
\end{table}

\begin{table}[t]
\centering
\caption{\scriptsize \label{tab:celeb}\textbf{Comparisons results on Celeb-reID.} ``pose'' represents human poses. Methods marked with ``$\dag$'' adopt DenseNet-121 as the backbone.}
\scriptsize
\setlength{\tabcolsep}{3.8mm}{
\begin{tabular}{llccc}
\toprule
\textbf{Methods} & \textbf{Modality} & \textbf{Rank-1} & \textbf{Rank-5} & \textbf{mAP}  \\ 
\midrule
PCB~\cite{sun2018beyond} & RGB & 37.1   & 57.0   & 8.2  \\
MGN~\cite{wang2018learning} & RGB & 49.0   & 64.9   & 10.8 \\ 
CESD~\cite{qian2020long} & RGB+pose & 50.9   & 66.3   & 9.8  \\
ReIDCaps~\cite{huang2019beyond} & RGB & 51.2   & 65.4   & 9.8  \\
IS-GAN$_{KL}$~\cite{eom2021disentangled} & RGB & 54.5   & -      & 12.8 \\ 
RCSANet$^{\dag}$~\cite{huang2021clothing} & RGB & 55.6   & -      & 11.9 \\ 
SirNet$^{\dag}$~\cite{yang2022sampling} & RGB & 56.0 & 70.3 & 14.2 \\ 
CAMC~\cite{wang2022co} & RGB+pose & 57.5 & 71.5 & 12.3 \\ 
\midrule
CSSC (Ours)  & RGB  & \textbf{64.5}   & \textbf{78.1}   & \textbf{17.3}  \\ \bottomrule
\end{tabular}}
\end{table}

\subsection{Comparison with State-of-the-Art Methods}
\label{subsec:comparison}
\noindent \textbf{Results on PRCC and LTCC.}
In Tab.~\ref{tab:prcc_ltcc}, we compare with advanced methods designed for conventional person Re-ID~\cite{li2018harmonious,sun2018beyond,hou2019interaction,he2021transreid} and methods tailored for cloth-changing person Re-ID using auxiliary modalities~\cite{hong2021fine,jin2022cloth,wang2022co} and using DG-Net~\cite{zheng2019joint,yang2023win}.
CSSC with only RGB modality shows significant superiority over them, without relying on extra annotations or auxiliary modalities.
Recently, some works have proposed to utilize ground-truth clothing labels to mitigate the impact of clothes~\cite{gu2022clothes,yang2023good,han2023clothing},
and some also use auxiliary modalities~\cite{chen2021learning,yan2022weakening,cui2023dcr}. 
CSSC achieves state-of-the-art results on PRCC and exhibits competitive performance with them on LTCC.
It is worth noting that some advanced methods~\cite{huang2021clothing,gu2022clothes,yang2023good,cui2023dcr} also incorporate both average and max poolings, but they simply concatenate the two pooled features at the end. The advantages of both poolings are not exerted.
However, CSSC using abundant semantics shows great advantages over them, without relying on extra annotations or auxiliary information. 
Our SMR module can be seamlessly integrated into the backbone without imposing a significant computational burden. The total number of parameters on LTCC is 54.3M for CSSC, while 141.2M for AIM~\cite{yang2023good} and 62.0M for CAMC~\cite{wang2022co}. The training FLOPs on LTCC are 12.2G for CSSC, while 18.5G for AIM~\cite{yang2023good} and 19.7G for CAMC~\cite{wang2022co}.

\noindent \textbf{Results on Celeb-reID.}
As shown in Tab.~\ref{tab:celeb}, all competitors achieve relatively poor performance, even those using stronger DenseNet-121~\cite{huang2017densely,huang2021clothing,yang2022sampling} as the backbone.
Despite the utilization of other modalities by advanced methods~\cite{qian2020long,wang2022co}, CSSC excels them without bells and whistles. 
It is worth noting that methods relying on manually annotated clothing labels, such as CAL~\cite{gu2022clothes}, AIM~\cite{yang2023good}, and CCFA~\cite{han2023clothing}, cannot work on Celeb-reID since the clothing annotations are not available. In contrast, our CSSC achieves remarkable performance without relying on extra impractical annotations.

\begin{table}[t]
\centering
\caption{\scriptsize \label{tab:ablation_SMR}\textbf{Ablation studies of the SMR module.} We report the results on PRCC and LTCC under the cloth-changing setting. ``Local.'' and ``Refine.'' denote local semantics mining and semantics refinement in SMR.
}
\scriptsize
\setlength{\tabcolsep}{4.5mm}{
\begin{tabular}{lcccc}
\toprule
{\multirow{2}{*}{\textbf{Methods}}} & \multicolumn{2}{c}{\textbf{PRCC}} & \multicolumn{2}{c}{\textbf{LTCC}}                    \\ \cmidrule(r){2-3} \cmidrule(r){4-5}
     & Rank-1 & mAP & Rank-1 & mAP \\ \midrule
Ours \textit{w/o} SMR     & 57.8 & 55.8 & 39.3 & 16.4 \\
Ours \textit{w/o} Local.  & 63.1 & 60.5 & 42.1 & 17.7 \\
Ours \textit{w/o} Refine. & 62.8 & 60.2 & 42.1 & 18.2 \\
\midrule
Ours                      & \textbf{65.5} & \textbf{63.0} & \textbf{43.6} & \textbf{18.6} \\ \bottomrule
\end{tabular}}
\end{table}

\subsection{Ablation Studies}
\label{subsec:ablation}
\noindent \textbf{Effectiveness of the SMR module.}
As shown in Tab.~\ref{tab:ablation_SMR}, there is a substantial performance decrease without using our SMR module. The results show the undeniable importance of SMR.
Additionally, both local semantics mining and semantics refinement designs contribute to performance improvement, further demonstrating the effectiveness of our proposed SMR.

\begin{table}
\centering
\caption{\scriptsize \label{tab:ablation_framework}\textbf{Ablation of the cross-parallel semantics collaboration framework.} We report the results on PRCC and LTCC under the cloth-changing setting.
``SMR-C-S (SMR-S-C)" denotes sequentially using SMR-C and SMR-S (SMR-S and SMR-C) in one branch.
}
\resizebox{1\linewidth}{!}{
\begin{tabular}{ccccccccccc}
\toprule
{\multirow{2}{*}{\textbf{Methods}}} & \multicolumn{4}{c}{\textbf{Branch 1}} & \multicolumn{2}{c}{\textbf{Branch 2}} & \multicolumn{2}{c}{\textbf{PRCC}} & \multicolumn{2}{c}{\textbf{LTCC}} \\
\cmidrule(r){2-5} \cmidrule(r){6-7} \cmidrule(r){8-9} \cmidrule(r){10-11}
\multicolumn{1}{c}{} & SMR-C & SMR-S & SMR-C-S & SMR-S-C & SMR-S & SMR-S-C & Rank-1 & mAP & Rank-1 & mAP \\
\midrule
1 & & & & & & & 57.8 & 55.8 & 39.3 & 16.4 \\
2 & $\checkmark$ &  &  &  &  &  & 61.7 & 58.3 & 40.6 & 17.1 \\
3 &  & $\checkmark$ &  &  &  &  & 60.3 & 58.8 & 41.1 & 17.8 \\
4 & $\checkmark$ &  &  &  & $\checkmark$ &  & 63.7 & 60.3 & 41.6 & 18.1 \\
5 &  &  & $\checkmark$ &  &  &  & 62.0 & 58.7 & 41.8 & 17.4 \\
6 &  &  &  & $\checkmark$ &  &  & 63.6 & 60.9 & 41.3 & 18.2 \\
\midrule
Ours &  &  & $\checkmark$ &  &  & $\checkmark$ & \textbf{65.5} & \textbf{63.0} & \textbf{43.6} & \textbf{18.6} \\
\bottomrule
\end{tabular}}
\end{table}

\noindent \textbf{Effectiveness of our cross-parallel semantics collaboration design.} 
To demonstrate that the great success of our method comes from our semantics collaboration design, rather than simply introducing multiple branches or several Re-ID losses, we ablate each branch by trying different combinations of SMR in Tab.~\ref{tab:ablation_framework}.
Either content or salient semantics (Methods 2 and 3) can improve the capability of the Re-ID model, but only one kind of semantics is suboptimal. Collaborating the two semantics in parallel (Method 4) or serial (Methods 5 and 6) bring limited improvement, while our cross-parallel collaboration framework exerts the potential of both semantics.

\begin{figure}[t]
\centering
  \includegraphics[width=0.9\linewidth]{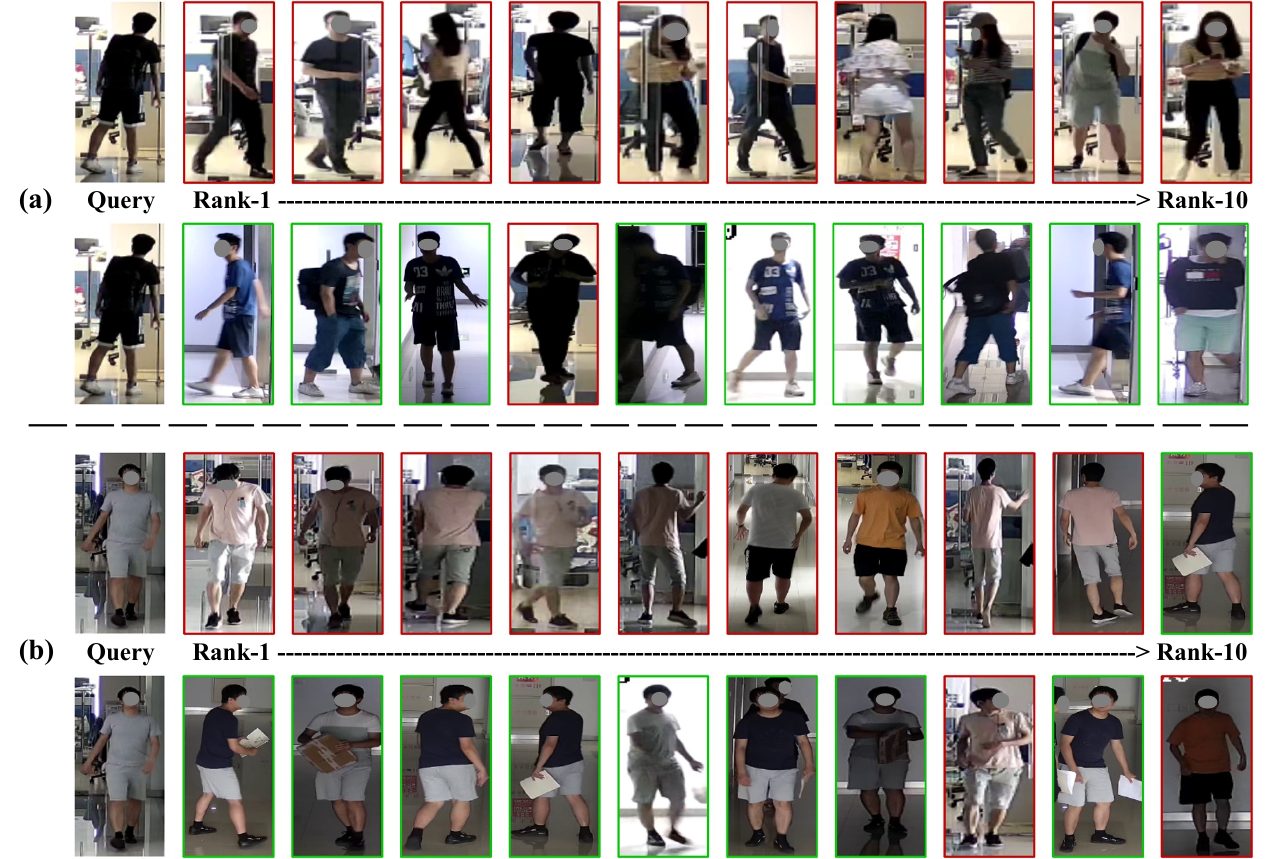}
  \caption{\textbf{Visualization of top-10 retrieval results.} For each query image, the first and the second rows are the ordered matching results obtained by using the baseline ResNet-50 and our proposed CSSC, respectively. Images with green and red borders indicate correct and wrong matching results, respectively. The results are obtained in the cloth-changing setting on LTCC. 
  }
  \vspace{-0.1in}
  \label{fig:retrieval}
\end{figure}

\noindent \textbf{Visualization of retrieval results.}
To intuitively demonstrate the effect of abundant semantics in our proposed method to promote cloth-changing person Re-ID, we compare the retrieval results of CSSC and the baseline model in Fig.~\ref{fig:retrieval}. As shown in the first-row results of (a) and (b), when people change their clothes, the baseline model is unable to identify pedestrians correctly with the interference of similar visual appearances, \textit{e.g.}, similar colors, and clothing textures. However, the pedestrian images with different clothes are correctly retrieved as the top results for our method. As observed from the second-row results of (a) and (b), identity-related semantics in global and local like body and part shapes excavated by our proposed SMR module help re-identify the same pedestrian despite changing clothes. Our proposed CSSC is better at capturing content and salient semantics, such as discriminative shorts and shoes in (b), to get good retrieval results.

\section{Conclusion}
In this paper, we address the cloth-changing challenge and propose a novel Content and Salient Semantics Collaboration (CSSC) framework with effective semantic interaction. 
In the framework, we propose the Semantics Mining and Refinement (SMR) module equipped with distinct pooling strategies to mine and leverage identity-related content semantics and identity-related salient semantics.
Extensive experiments demonstrate the effectiveness and superiority of our method. 
We hope this paper can inspire more research to advance cloth-changing person Re-ID without relying on unpractical and inflexible auxiliary modalities or extra annotations.



\bibliographystyle{IEEEtran}
\bibliography{egbib}

\begin{thebibliography}{10}
\providecommand{\url}[1]{#1}
\csname url@samestyle\endcsname
\providecommand{\newblock}{\relax}
\providecommand{\bibinfo}[2]{#2}
\providecommand{\BIBentrySTDinterwordspacing}{\spaceskip=0pt\relax}
\providecommand{\BIBentryALTinterwordstretchfactor}{4}
\providecommand{\BIBentryALTinterwordspacing}{\spaceskip=\fontdimen2\font plus
\BIBentryALTinterwordstretchfactor\fontdimen3\font minus \fontdimen4\font\relax}
\providecommand{\BIBforeignlanguage}[2]{{%
\expandafter\ifx\csname l@#1\endcsname\relax
\typeout{** WARNING: IEEEtran.bst: No hyphenation pattern has been}%
\typeout{** loaded for the language `#1'. Using the pattern for}%
\typeout{** the default language instead.}%
\else
\language=\csname l@#1\endcsname
\fi
#2}}
\providecommand{\BIBdecl}{\relax}
\BIBdecl

\bibitem{qian2017multi}
X.~Qian, Y.~Fu, Y.-G. Jiang, T.~Xiang, and X.~Xue, ``Multi-scale deep learning architectures for person re-identification,'' in \emph{Proceedings of the IEEE international conference on computer vision}, 2017, pp. 5399--5408.

\bibitem{hou2019interaction}
R.~Hou, B.~Ma, H.~Chang, X.~Gu, S.~Shan, and X.~Chen, ``Interaction-and-aggregation network for person re-identification,'' in \emph{Proceedings of the IEEE/CVF Conference on Computer Vision and Pattern Recognition}, 2019, pp. 9317--9326.

\bibitem{sun2018beyond}
Y.~Sun, L.~Zheng, Y.~Yang, Q.~Tian, and S.~Wang, ``Beyond part models: Person retrieval with refined part pooling (and a strong convolutional baseline),'' in \emph{Proceedings of the European Conference on Computer Vision}, 2018, pp. 480--496.

\bibitem{wang2023rethinking}
Q.~Wang, X.~Qian, B.~Li, Y.~Fu, and X.~Xue, ``Rethinking person re-identification from a projection-on-prototypes perspective,'' \emph{arXiv preprint arXiv:2308.10717}, 2023.

\bibitem{meng2024unleashing}
N.~Meng, Q.~Wang, B.~Li, and X.~Xue, ``Unleashing the potential of tracklets for unsupervised video person re-identification,'' \emph{arXiv preprint arXiv:2406.14261}, 2024.

\bibitem{wang2024distribution}
Q.~Wang, X.~Qian, B.~Li, and X.~Xue, ``Distribution aligned semantics adaption for lifelong person re-identification,'' \emph{arXiv preprint arXiv:2405.19695}, 2024.

\bibitem{wang2024large}
Q.~Wang, B.~Li, and X.~Xue, ``When large vision-language models meet person re-identification,'' \emph{arXiv preprint arXiv:2411.18111}, 2024.

\bibitem{qian2020long}
X.~Qian, W.~Wang, L.~Zhang, F.~Zhu, Y.~Fu, T.~Xiang, Y.-G. Jiang, and X.~Xue, ``Long-term cloth-changing person re-identification,'' in \emph{Proceedings of the Asian Conference on Computer Vision}, 2020, pp. 71--88.

\bibitem{yang2019person}
Q.~Yang, A.~Wu, and W.-S. Zheng, ``Person re-identification by contour sketch under moderate clothing change,'' \emph{IEEE Transactions on Pattern Analysis and Machine Intelligence}, vol.~43, no.~6, pp. 2029--2046, 2019.

\bibitem{wang2022co}
Q.~Wang, X.~Qian, Y.~Fu, and X.~Xue, ``Co-attention aligned mutual cross-attention for cloth-changing person re-identification,'' in \emph{Proceedings of the Asian Conference on Computer Vision}, 2022, pp. 2270--2288.

\bibitem{wang2024image}
Q.~Wang, X.~Qian, B.~Li, Y.~Fu, and X.~Xue, ``Image-text-image knowledge transferring for lifelong person re-identification with hybrid clothing states,'' \emph{arXiv preprint arXiv:2405.16600}, 2024.

\bibitem{hong2021fine}
P.~Hong, T.~Wu, A.~Wu, X.~Han, and W.-S. Zheng, ``Fine-grained shape-appearance mutual learning for cloth-changing person re-identification,'' in \emph{Proceedings of the IEEE/CVF Conference on Computer Vision and Pattern Recognition}, 2021, pp. 10\,513--10\,522.

\bibitem{jin2022cloth}
X.~Jin, T.~He, K.~Zheng, Z.~Yin, X.~Shen, Z.~Huang, R.~Feng, J.~Huang, Z.~Chen, and X.-S. Hua, ``Cloth-changing person re-identification from a single image with gait prediction and regularization,'' in \emph{Proceedings of the IEEE/CVF Conference on Computer Vision and Pattern Recognition}, 2022, pp. 14\,278--14\,287.

\bibitem{gu2022clothes}
X.~Gu, H.~Chang, B.~Ma, S.~Bai, S.~Shan, and X.~Chen, ``Clothes-changing person re-identification with rgb modality only,'' in \emph{Proceedings of the IEEE/CVF Conference on Computer Vision and Pattern Recognition}, 2022, pp. 1060--1069.

\bibitem{yang2023good}
Z.~Yang, M.~Lin, X.~Zhong, Y.~Wu, and Z.~Wang, ``Good is bad: Causality inspired cloth-debiasing for cloth-changing person re-identification,'' in \emph{Proceedings of the IEEE/CVF Conference on Computer Vision and Pattern Recognition}, 2023, pp. 1472--1481.

\bibitem{han2023clothing}
K.~Han, S.~Gong, Y.~Huang, L.~Wang, and T.~Tan, ``Clothing-change feature augmentation for person re-identification,'' in \emph{Proceedings of the IEEE/CVF Conference on Computer Vision and Pattern Recognition}, 2023, pp. 22\,066--22\,075.

\bibitem{wang2018learning}
G.~Wang, Y.~Yuan, X.~Chen, J.~Li, and X.~Zhou, ``Learning discriminative features with multiple granularities for person re-identification,'' in \emph{Proceedings of the 26th ACM International Conference on Multimedia}, 2018, pp. 274--282.

\bibitem{luo2019bag}
H.~Luo, Y.~Gu, X.~Liao, S.~Lai, and W.~Jiang, ``Bag of tricks and a strong baseline for deep person re-identification,'' in \emph{Proceedings of the IEEE/CVF Conference on Computer Vision and Pattern Recognition Workshops}, 2019, pp. 0--0.

\bibitem{hermans2017defense}
A.~Hermans, L.~Beyer, and B.~Leibe, ``In defense of the triplet loss for person re-identification,'' \emph{arXiv preprint arXiv:1703.07737}, 2017.

\bibitem{hu2018squeeze}
J.~Hu, L.~Shen, and G.~Sun, ``Squeeze-and-excitation networks,'' in \emph{Proceedings of the IEEE/CVF Conference on Computer Vision and Pattern Recognition}, 2018, pp. 7132--7141.

\bibitem{huang2019beyond}
Y.~Huang, J.~Xu, Q.~Wu, Y.~Zhong, P.~Zhang, and Z.~Zhang, ``Beyond scalar neuron: Adopting vector-neuron capsules for long-term person re-identification,'' \emph{IEEE Transactions on Circuits and Systems for Video Technology}, vol.~30, no.~10, pp. 3459--3471, 2019.

\bibitem{resnet}
K.~He, X.~Zhang, S.~Ren, and J.~Sun, ``Deep residual learning for image recognition,'' in \emph{Proceedings of the IEEE/CVF Conference on Computer Vision and Pattern Recognition}, 2016, pp. 770--778.

\bibitem{deng2009imagenet}
J.~Deng, W.~Dong, R.~Socher, L.-J. Li, K.~Li, and L.~Fei-Fei, ``Imagenet: A large-scale hierarchical image database,'' in \emph{Proceedings of the IEEE/CVF Conference on Computer Vision and Pattern Recognition}, 2009, pp. 248--255.

\bibitem{cui2023dcr}
Z.~Cui, J.~Zhou, Y.~Peng, S.~Zhang, and Y.~Wang, ``Dcr-reid: Deep component reconstruction for cloth-changing person re-identification,'' \emph{IEEE Transactions on Circuits and Systems for Video Technology}, 2023.

\bibitem{yan2022weakening}
Y.~Yan, H.~Yu, S.~Li, Z.~Lu, J.~He, H.~Zhang, and R.~Wang, ``Weakening the influence of clothing: Universal clothing attribute disentanglement for person re-identification,'' in \emph{Proceedings of the International Joint Conference on Artificial Intelligence}, 2022, pp. 1523--1529.

\bibitem{zhong2020random}
Z.~Zhong, L.~Zheng, G.~Kang, S.~Li, and Y.~Yang, ``Random erasing data augmentation,'' in \emph{Proceedings of the AAAI Conference on Artificial Intelligence}, vol.~34, no.~07, 2020, pp. 13\,001--13\,008.

\bibitem{kingma2014adam}
D.~P. Kingma and J.~Ba, ``Adam: A method for stochastic optimization,'' in \emph{Proceedings of the International Conference on Learning Representations}, 2015.

\bibitem{wang2024exploring}
Q.~Wang, X.~Qian, B.~Li, X.~Xue, and Y.~Fu, ``Exploring fine-grained representation and recomposition for cloth-changing person re-identification,'' \emph{IEEE Transactions on Information Forensics and Security}, 2024.

\bibitem{li2018harmonious}
W.~Li, X.~Zhu, and S.~Gong, ``Harmonious attention network for person re-identification,'' in \emph{Proceedings of the IEEE/CVF Conference on Computer Vision and Pattern Recognition}, 2018, pp. 2285--2294.

\bibitem{he2021transreid}
S.~He, H.~Luo, P.~Wang, F.~Wang, H.~Li, and W.~Jiang, ``Transreid: Transformer-based object re-identification,'' in \emph{Proceedings of the IEEE/CVF International Conference on Computer Vision}, 2021, pp. 15\,013--15\,022.

\bibitem{huang2021clothing}
Y.~Huang, Q.~Wu, J.~Xu, Y.~Zhong, and Z.~Zhang, ``Clothing status awareness for long-term person re-identification,'' in \emph{Proceedings of the IEEE/CVF International Conference on Computer Vision}, 2021, pp. 11\,895--11\,904.

\bibitem{yang2023win}
Z.~Yang, X.~Zhong, Z.~Zhong, H.~Liu, Z.~Wang, and S.~Satoh, ``Win-win by competition: Auxiliary-free cloth-changing person re-identification,'' \emph{IEEE Transactions on Image Processing}, vol.~32, pp. 2985--2999, 2023.

\bibitem{eom2021disentangled}
C.~Eom, W.~Lee, G.~Lee, and B.~Ham, ``Disentangled representations for short-term and long-term person re-identification,'' \emph{IEEE Transactions on Pattern Analysis and Machine Intelligence}, vol.~44, no.~12, pp. 8975--8991, 2021.

\bibitem{yang2022sampling}
S.~Yang, B.~Kang, and Y.~Lee, ``Sampling agnostic feature representation for long-term person re-identification,'' \emph{IEEE Transactions on Image Processing}, vol.~31, pp. 6412--6423, 2022.

\bibitem{zheng2019joint}
Z.~Zheng, X.~Yang, Z.~Yu, L.~Zheng, Y.~Yang, and J.~Kautz, ``Joint discriminative and generative learning for person re-identification,'' in \emph{Proceedings of the IEEE/CVF Conference on Computer Vision and Pattern Recognition}, 2019, pp. 2138--2147.

\bibitem{chen2021learning}
J.~Chen, X.~Jiang, F.~Wang, J.~Zhang, F.~Zheng, X.~Sun, and W.-S. Zheng, ``Learning 3d shape feature for texture-insensitive person re-identification,'' in \emph{Proceedings of the IEEE/CVF Conference on Computer Vision and Pattern Recognition}, 2021, pp. 8146--8155.

\bibitem{huang2017densely}
G.~Huang, Z.~Liu, K.~Q. Weinberger, and L.~van~der Maaten, ``Densely connected convolutional networks,'' in \emph{Proceedings of the IEEE/CVF Conference on Computer Vision and Pattern Recognition}, vol.~1, no.~2, 2017, p.~3.

\end{thebibliography}

\end{document}